\documentclass[conference]{IEEEtran}

\usepackage{cite}
\usepackage{amsmath,amssymb,amsfonts}
\usepackage{algorithmic}
\usepackage{graphicx}
\usepackage{textcomp}
\usepackage{xcolor}
\usepackage{multirow}
\usepackage{booktabs}
\usepackage{colortbl}
\usepackage{float}
\usepackage{pifont}
\usepackage[hidelinks]{hyperref}
\newcommand{\cmark}{\ding{51}} 

\def\BibTeX{{\rm B\kern-.05em{\sc i\kern-.025em b}\kern-.08em
    T\kern-.1667em\lower.7ex\hbox{E}\kern-.125emX}}
\begin{document}

\title{Learning Hierarchical Orthogonal Prototypes for Generalized Few-Shot 3D Point Cloud Segmentation}

\author{
    \IEEEauthorblockN{
        Yifei Zhao\textsuperscript{$\ast$}, 
        Fanyu Zhao\textsuperscript{$\ast$}, 
        Zhongyuan Zhang, 
        Shengtang Wu, 
        Yixuan Lin, 
        and Yinsheng Li\textsuperscript{$\dagger$}
    }
    \IEEEauthorblockA{College of Computer Science and Artificial Intelligence, Fudan University, Shanghai, China \\
    Email: \{yfzhao19, fyzhao20, liys\}@fudan.edu.cn, \{zhongyuanzhang25, ctwu25, yixuanlin24\}@m.fudan.edu.cn}
    \vspace{-25pt}
    \thanks{$\ast$ Equal contribution. $\dagger$ Corresponding author.}
}
\IEEEoverridecommandlockouts
\maketitle

\begin{abstract}
Generalized few-shot 3D point cloud segmentation aims to adapt to novel classes from only a few annotations while maintaining strong performance on base classes, but this remains challenging due to the inherent stability--plasticity trade-off: adapting to novel classes can interfere with shared representations and cause base-class forgetting.
We present HOP3D, a unified framework that learns hierarchical orthogonal prototypes with an entropy-based few-shot regularizer to enable robust novel-class adaptation without degrading base-class performance.
HOP3D introduces hierarchical orthogonalization that decouples base and novel learning at both the gradient and representation levels, effectively mitigating base--novel interference.
To further enhance adaptation under sparse supervision, we incorporate an entropy-based regularizer that leverages predictive uncertainty to refine prototype learning and promote balanced predictions.
Extensive experiments on ScanNet200 and ScanNet++ demonstrate that HOP3D consistently outperforms state-of-the-art baselines under both 1-shot and 5-shot settings. The code is available at \href{https://fdueblab-hop3d.github.io/}{\textbf{fdueblab-hop3d.github.io}}.
\end{abstract}


\section{Introduction}
3D point cloud semantic segmentation assigns a label to each point in a 3D scene and underpins autonomous driving, robotics, and AR/VR applications~\cite{dai2017scannet, qi2017pointnet++, thomas2019KPConv}.
With large-scale dense annotations, fully supervised models have achieved remarkable progress~\cite{schult2023mask3d, kolodiazhnyi2024oneformer3d, lai2022stratified, han2024asgformer}, yet high-quality 3D annotation is costly and difficult to scale~\cite{sun2024review, wang2024survey}.
This motivates few-shot 3D segmentation, which adapts to novel categories from only a few labeled examples~\cite{he2023prototype, an2024rethinking, xu2024partwhole, an2025mmfss, zhao2026upl, wei2025few}.

A more realistic setting is generalized few-shot 3D point cloud segmentation (GFS-3DS), where the model must recognize base classes with abundant supervision and novel classes with sparse annotations simultaneously~\cite{xu2023generalized, an2025generalized}.
GFS-3DS is fundamentally constrained by the stability--plasticity dilemma: improving novel-class performance often degrades base-class knowledge~\cite{kirkpatrick2017ewc, parisi2019continual, kim2023ancl}.
This conflict is particularly acute for prototype-based formulations widely used in few-/generalized few-shot 3D segmentation~\cite{zhao2021few, an2024rethinking, xu2023generalized, an2025generalized, yang2024gfs3d}.
First, base and novel categories share the same feature space and parameters, so few-shot updates for novel classes can directly perturb base decision boundaries.
Second, segmentation is prototype-centric: predictions are governed by the relative geometry between point embeddings and class prototypes.
Under sparse and biased support, novel prototypes are often noisy; updating them can warp the prototype subspace structure, making base--novel separation fragile and amplifying interference~\cite{qi2017pointnet++, xu2023generalized, an2025generalized}.
In short, reducing interference requires controlling both (i) the few-shot adaptation dynamics (\emph{how to learn}) and (ii) the prototype subspace structure that shapes the decision geometry (\emph{what to learn}).

Orthogonality provides a natural guiding principle for addressing both levels~\cite{farajtabar2020orthogonal,liu2023learning,chowdhury2022fscil3d,bansal2018orthogonal,cogswell2015decov}.
Prior work shows that orthogonal gradient projection can mitigate forgetting in continual learning~\cite{farajtabar2020orthogonal} and that orthogonal prototypes improve base--novel separation in generalized few-shot \emph{2D} segmentation~\cite{liu2023learning}; related orthogonal-basis designs have also appeared in 3D class-incremental learning~\cite{chowdhury2022fscil3d}.
However, orthogonality must be enforced at \emph{both} levels to be effective in GFS-3DS:
gradient-space orthogonalization alone can stabilize updates but cannot prevent few-shot noise from geometrically warping the prototype subspace,
while prototype-space orthogonality alone improves separability yet cannot stop novel adaptation from perturbing shared parameters and forgetting base knowledge.
Therefore, a joint design that aligns projected updates with prototype subspace geometry is necessary for reliable base--novel joint recognition,
but such a two-level orthogonality coupling remains underexplored in GFS-3DS.

Motivated by this, we propose \textbf{HOP3D}, a unified framework that instantiates orthogonality at both the optimization level and the representation level, together with an entropy-aware few-shot regularizer.
Specifically, \textbf{HOP-Net} performs hierarchical orthogonalization via:
(i) \textbf{HOP-Grad}, which projects novel gradients onto the orthogonal complement of base gradient directions to suppress harmful interference during Phase~2 adaptation; and
(ii) \textbf{HOP-Rep}, which learns orthogonal prototype subspaces to induce a base/novel representation decomposition, improving separability while preserving base knowledge.
To further improve robustness under extremely limited supervision, we introduce \textbf{HOP-Ent}, a dual-entropy regularizer (conditional-entropy minimization and marginal-entropy maximization) integrated into few-shot training to sharpen and balance novel predictions, avoiding extra test-time optimization ~\cite{hajimiri2023strong}.
Our main contributions are:
\begin{itemize}
    \item From a unified view (\emph{how vs.\ what to learn}) of GFS-3DS, we propose \textbf{HOP-Net}, which instantiates a joint orthogonality principle via gradient-space orthogonal projection (\textbf{HOP-Grad}) and prototype-space orthogonal decomposition (\textbf{HOP-Rep}) to mitigate base--novel interference.
    \item We introduce \textbf{HOP-Ent}, a dual-entropy regularizer integrated into Phase~2 training to improve prediction certainty and class balance.
    \item Extensive experiments on \textbf{ScanNet200} and \textbf{ScanNet++} demonstrate that \textbf{HOP3D} achieves state-of-the-art performance under both 1-shot and 5-shot settings.
\end{itemize}

\begin{figure*}[t]
    \centering
    \includegraphics[width=0.9\linewidth]{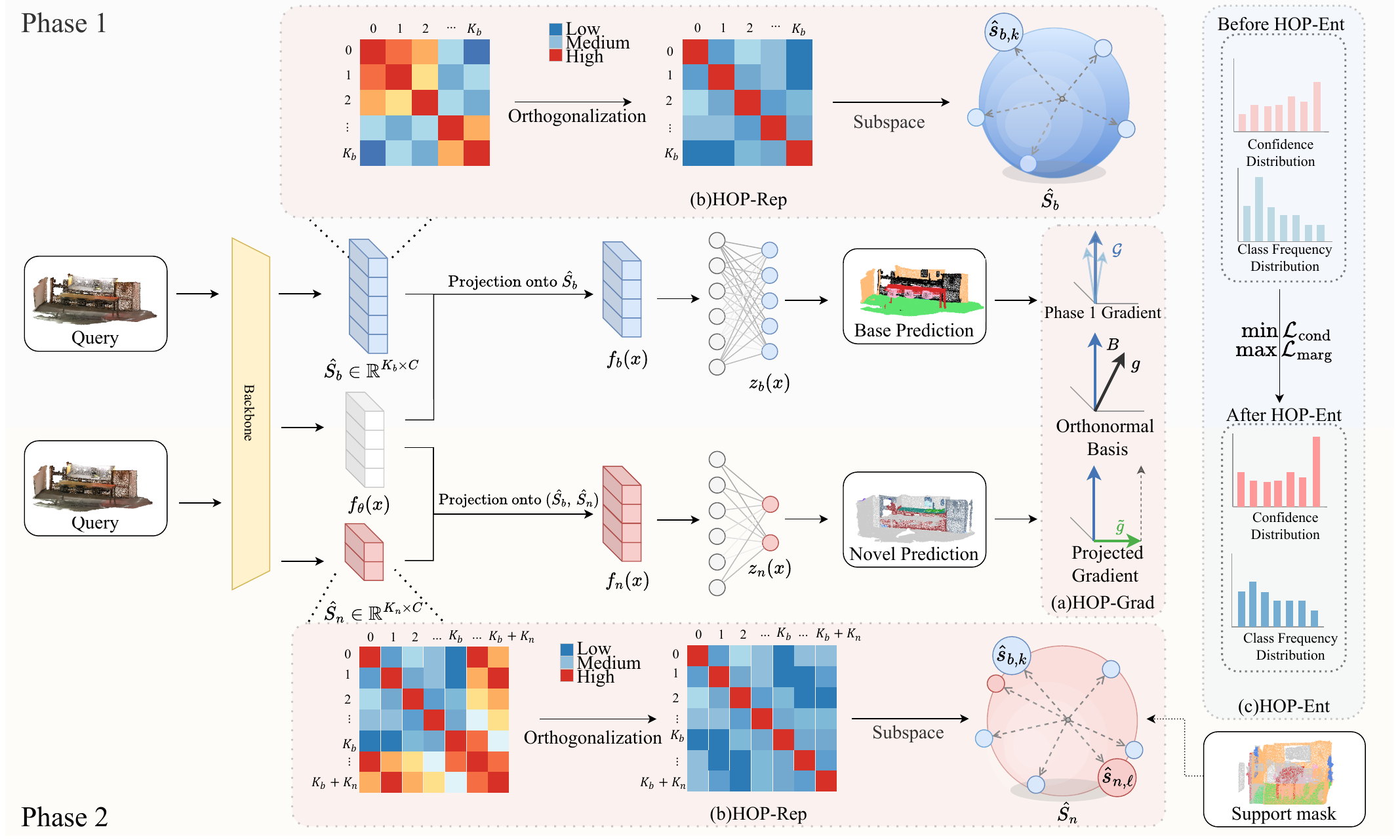}
    \vspace{-5pt}
    \caption{Overview of the proposed HOP3D framework, integrating HOP-Net and HOP-Ent.
    The training pipeline consists of two phases:
    \textbf{Phase~1} trains base classes with \textbf{HOP-Rep} and collects base-task gradients $\mathcal{G}$ to construct an orthonormal basis $B$ for \textbf{HOP-Grad};
    \textbf{Phase~2} introduces novel classes for few-shot adaptation, applies \textbf{HOP-Grad} to project each novel gradient $g$ as $\tilde{g}$ (removing $B$), continues to use \textbf{HOP-Rep} for prototype subspace orthogonalization, and integrates \textbf{HOP-Ent} for entropy-guided refinement.
    (a) Gradient orthogonalization in \textbf{HOP-Grad} using $B$ and $\tilde g$;
    (b) Prototype similarity heatmaps and orthogonal prototype subspaces before/after \textbf{HOP-Rep};
    (c) Confidence distribution and class-frequency distribution before/after \textbf{HOP-Ent}.}
    \vspace{-8pt}
    \label{fig:framework}
\end{figure*}

\section{Methodology}
\subsection{Overview}
We propose HOP3D, a unified framework for GFS-3DS that reduces base–novel interference and improves generalization under limited supervision. As shown in Fig.~\ref{fig:framework}, HOP3D integrates two complementary components:
(1) HOP-Net, which performs hierarchical orthogonalization, applying gradient-level orthogonal projection to avoid base-class forgetting and representation-level prototype orthogonalization to strengthen semantic separation; and (2) HOP-Ent, an entropy-based few-shot regularizer that encourages confident and balanced predictions. HOP3D is trained in two phases: base pretraining and novel adaptation. During the latter, both HOP-Net and HOP-Ent are activated to jointly enhance optimization stability and few-shot generalization.

\subsection{Hierarchical Orthogonal Prototype Network (HOP-Net)}
\noindent\textbf{Orthogonal Gradient Projection (HOP-Grad):}
At the gradient level, HOP-Net incorporates an orthogonal projection module to stabilize few-shot adaptation while preserving base-class knowledge. Inspired by continual learning, this module prevents gradients from novel-class samples from updating directions already optimized for base classes by projecting them onto the orthogonal complement of the base gradient subspace.
Let $\phi \in \mathbb{R}^d$ denote segmentation parameters, including class prototypes and classifier weights, where $d$ is the dimensionality of the vectorized parameter set.

Upon completing Phase~1 training, we extract a representative set of gradients $\mathcal{G} = \{g^{(t)}\}_{t=1}^T$ from the converged model by re-processing $T$ mini-batches from the base training set, where $T$ is chosen to balance gradient diversity and computational cost. 
Each $g^{(t)} \in \mathbb{R}^d$ denotes the gradient of the Phase~1 objective with respect to $\phi$ computed at the $t$-th extraction step.
We then apply the Gram--Schmidt process to $\mathcal{G}$ to construct a compact orthonormal basis $B \in \mathbb{R}^{d \times r}$, where $r \leq T$ corresponds to the effective rank of the gradient set:
\begin{equation}
B = \text{GS}(\{g^{(t)}\}_{t=1}^T), \quad \text{with} \quad B^\top B = I.
\end{equation}
The basis $B$ remains fixed throughout Phase~2, constraining gradient updates to lie within the orthogonal complement of the base optimization subspace.

During Phase~2, for any gradient $g \in \mathbb{R}^d$ with respect to $\phi$ that is induced by the novel-class supervision, we project it onto the orthogonal complement of the base subspace spanned by $B$ as $\tilde{g} = g - B(B^\top g)$. Since $B$ is orthonormal, $BB^\top g$ is the component of $g$ within $B$, and $\tilde g$ retains only the component orthogonal to $B$. This removes update directions that overlap with the base optimization subspace, helping mitigate base-class forgetting.

The final projected gradient $\tilde{g}$ is then used to update parameters $\phi$ via the chosen optimizer. This gradient-level decoupling forms the first component of HOP-Net, ensuring stable adaptation by explicitly separating base and novel update directions.

\noindent\textbf{Orthogonal Representation Decomposition (HOP-Rep):}
To disentangle representations, we enforce pairwise orthogonality on the parameterized projection bases, rather than raw features. This ensures both intra-group separation and inter-group independence between base ($\hat{S}_b$) and novel ($\hat{S}_n$) subspaces, enabling sequential projections to resolve features into mutually decorrelated semantic components.
Let $x \in \mathbb{R}^3$ denote a point in the input cloud, and let $f_\theta(x) \in \mathbb{R}^C$ be its embedded feature produced by the backbone $f_\theta(\cdot)$, where $C$ is the feature dimension. Define the $\ell_2$-normalized base prototype set for the $K_b$ base classes as $\hat{S}_b = \{\hat{s}_{b,k} \in \mathbb{R}^C\}_{k=1}^{K_b}$.

In Phase~1, only base prototypes are used. The input feature is first projected onto the subspace spanned by base prototypes:
\begin{equation}
f_b(x) = \sum_{k=1}^{K_b} \langle f_\theta(x), \hat{s}_{b,k} \rangle \hat{s}_{b,k},
\end{equation}
\begin{equation}
r^{(0)}(x) = f_\theta(x) - f_b(x),
\end{equation}
where $f_b(x) \in \mathbb{R}^C$ denotes the base-aligned component and $r^{(0)}(x) \in \mathbb{R}^C$ is the residual orthogonal to the base subspace. These two components are concatenated and passed through a shared multi-layer perceptron (MLP) to produce per-point classification scores for base classes.

In Phase~2, we introduce the novel prototype set $\hat{S}_n = \{\hat{s}_{n,\ell} \in \mathbb{R}^C\}_{\ell=1}^{K_n}$ for the $K_n$ novel classes. The residual $r^{(0)}(x)$ is then projected onto the corresponding novel subspace:

\begin{equation}
f_n(x) = \sum_{\ell=1}^{K_n} \langle r^{(0)}(x), \hat{s}_{n,\ell} \rangle \hat{s}_{n,\ell},
\end{equation}
\begin{equation}
r^{(1)}(x) = r^{(0)}(x) - f_n(x),
\end{equation}
where $f_n(x) \in \mathbb{R}^C$ is the novel-aligned projection and $r^{(1)}(x) \in \mathbb{R}^C$ is the remaining residual. The model employs two separate MLPs: $h_b(\cdot)$ for base features and $h_n(\cdot)$ for novel and residual features. The final prediction is obtained as:
\begin{equation}
z(x) = \left[ h_b(f_b(x)),\quad h_n(f_n(x), r^{(1)}(x)) \right],
\end{equation}
where $z(x) \in \mathbb{R}^{K_b+K_n}$ denotes the logits over all foreground categories. Although base prototypes in Phase~2 are initialized from those learned in Phase~1, their role is not strictly identical. In Phase~1, the residual space captures background information. In Phase~2, however, the introduction of novel prototypes redefines the residual structure, prompting further adaptation of the base prototypes to align with the new decomposition.

To encourage decorrelation among all learned prototypes, we apply a unified orthogonality regularizer to the cosine similarity between all distinct prototype pairs. In Phase~1, this regularizer is applied only to base prototypes, and in Phase~2 it is extended to the joint set of base and novel prototypes:
\begin{equation}
\mathcal{L}_{\text{orth}} = \sum_{i < j} \big|\hat{s}_i^\top \hat{s}_j\big|,
\end{equation}
where the summation spans all distinct pairs of $\ell_2$-normalized prototypes in the current phase.

\subsection{Entropy-based Few-Shot Regularizer (HOP-Ent)}

To enhance generalization on novel categories under limited supervision, we introduce the Entropy-based Few-Shot Regularizer (HOP-Ent). This module jointly optimizes two complementary entropy-based objectives to encourage both confident and balanced predictions. 
During Phase~2, we generate novel-class supervision by adopting the pseudo-label selection and adaptive infilling of GFS-VL~\cite{an2025generalized} without any modification.

Let the model output logits for point $x$ be $z(x) \in \mathbb{R}^{K_b + K_n}$, and define $p(y \mid x) = \mathrm{softmax}(z(x))$ as the predicted class distribution, where $K_b$ and $K_n$ denote the number of base and novel classes. HOP-Ent consists of two loss components:

\noindent\textbf{Conditional Entropy Minimization.}  
For a selected set $\mathcal{S}$ of high-confidence pseudo-labeled points, we minimize the entropy of each prediction to improve per-sample certainty:
\begin{equation}
\mathcal{L}_{\mathrm{cond}} = \frac{1}{|\mathcal{S}|} \sum_{x \in \mathcal{S}} \left[ -\sum_{c \in \mathcal{C}_n} p(c \mid x) \log p(c \mid x) \right],
\end{equation}
where $\mathcal{C}_n$ is the set of novel class indices.

\noindent\textbf{Marginal Entropy Maximization.}  
To mitigate class imbalance among novel categories, we maximize the entropy of the batch-level marginal distribution:
\begin{equation}
\bar{p}(c) = \frac{1}{N} \sum_{i=1}^N p(c \mid x_i), \quad
\mathcal{L}_{\mathrm{marg}} = -\sum_{c \in \mathcal{C}_n} \bar{p}(c) \log \bar{p}(c),
\end{equation}
where $N$ is the total number of points in the batch.

\noindent\textbf{Total Regularization Loss.}  
The overall HOP-Ent loss is a weighted sum of the above terms:
\begin{equation}
\mathcal{L}_{\mathrm{ent}} = \lambda_{\mathrm{cond}} \, \mathcal{L}_{\mathrm{cond}} + \lambda_{\mathrm{marg}} \, \mathcal{L}_{\mathrm{marg}},
\end{equation}
where $\lambda_{\mathrm{cond}}$ and $\lambda_{\mathrm{marg}}$ control the trade-off between the two objectives.
Unlike test-time adaptation techniques, HOP-Ent is integrated into Phase~2 training and jointly updates all model parameters, including the backbone, classifier heads, and prototypes. This end-to-end optimization encourages confident and diverse predictions, significantly enhancing robustness in generalized few-shot settings.
\begin{table*}[t]
\caption{\textbf{Performance comparison on ScanNet200 and ScanNet++.} 
Results under 1-shot and 5-shot settings.}
\vspace{-6pt}
\centering
\renewcommand{\arraystretch}{1.05}
\resizebox{\linewidth}{!}{
\begin{tabular}{l|cccc|cccc||cccc|cccc}
\toprule
\multirow{3}{*}{Method} 
& \multicolumn{8}{c||}{ScanNet200} & \multicolumn{8}{c}{ScanNet++} \\
\cmidrule{2-9} \cmidrule{10-17}
& \multicolumn{4}{c|}{5-shot} & \multicolumn{4}{c||}{1-shot} 
& \multicolumn{4}{c|}{5-shot} & \multicolumn{4}{c}{1-shot} \\
\cmidrule{2-17}
& B & N & A & HM & B & N & A & HM & B & N & A & HM & B & N & A & HM \\
\midrule
Fully Sup. & 68.70 & 39.32 & 45.51 & 50.02 & 68.70 & 39.32 & 45.51 & 50.02 & 65.45 & 37.24 & 48.53 & 47.47 & 65.45 & 37.24 & 48.53 & 47.47 \\
PIFS~\cite{cermelli2021prototype} & 28.78 & 3.82 & 9.07 & 6.71 & 17.84 & 2.87 & 6.02 & 4.88 & 39.98 & 5.74 & 19.44 & 10.03 & 36.66 & 4.95 & 17.63 & 8.71 \\
attMPTI~\cite{zhao2021few} & 37.13 & 4.99 & 11.76 & 8.79 & 54.84 & 3.28 & 14.14 & 6.17 & 55.89 & 4.19 & 24.87 & 7.78 & 53.16 & 3.55 & 23.40 & 6.66 \\
COSeg~\cite{an2024rethinking} & 57.67 & 5.21 & 16.25 & 9.54 & 47.03 & 4.03 & 13.09 & 7.42 & 59.34 & 6.96 & 27.91 & 12.45 & 58.49 & 6.24 & 27.14 & 11.26 \\
GW~\cite{xu2023generalized} & 59.28 & 8.30 & 19.03 & 14.55 & 55.23 & 6.47 & 16.74 & 11.56 & 51.35 & 11.03 & 27.16 & 18.15 & 46.71 & 6.63 & 22.66 & 11.59 \\
GFS-VL~\cite{an2025generalized} & \textbf{67.57} & 31.67 & 39.23 & 43.12 & \textbf{68.48} & 29.18 & 37.45 & 40.92 & 60.05 & 21.66 & 37.02 & 31.82 & 61.39 & \textbf{19.42} & 36.21 & \textbf{29.47} \\
\rowcolor{gray!12}
\textbf{HOP3D (ours)} & 67.36 & \textbf{34.38} & \textbf{41.32} & \textbf{45.52} & 68.45 & \textbf{31.80} & \textbf{39.52} & \textbf{43.42} & \textbf{62.40} & \textbf{23.70} & \textbf{39.18} & \textbf{34.34} & \textbf{61.72} & 19.23 & \textbf{36.23} & 29.32 \\
\bottomrule
\end{tabular}
}
\vspace{-10pt}
\label{tab:main-combined}
\end{table*}

\subsection{Training Objective}
The overall training objective consists of two stages. In Phase~1, the model is trained on base classes using a segmentation loss and an orthogonality regularizer: 
\(
\mathcal{L}_{\text{P1}} = \mathcal{L}_{\text{seg}}^{\text{base}} + \lambda_{\text{orth}}^{(1)} \mathcal{L}_{\text{orth}}.
\) 
In Phase~2, the model adapts to novel classes using both labeled and pseudo-labeled data. The loss combines segmentation, dual entropy regularization, and continued orthogonality:
\(
\mathcal{L}_{\text{P2}} = \mathcal{L}_{\text{seg}} + \mathcal{L}_{\text{ent}} + \lambda_{\text{orth}}^{(2)} \mathcal{L}_{\text{orth}}.
\)
To prevent base forgetting, orthogonal gradient projection is applied during optimization:
\(
\tilde{g} = g - B(B^\top g)
\).

\section{Experiments}
\subsection{Experimental Setup}
\label{sec:expsetup}

\noindent\textbf{Datasets.} 
We evaluate our method on two large-scale benchmarks: \textbf{ScanNet200}~\cite{rozenberszki2022language}, an extension of ScanNet~\cite{dai2017scannet} to 200 categories, and \textbf{ScanNet++}~\cite{yeshwanth2023scannet++}, which comprises 460 scenes across over 1,000 unique classes. 
Following the GFS-PCS protocol~\cite{an2025generalized}, our benchmark incorporates 57 classes for ScanNet200 and 30 for ScanNet++, with official train/test splits maintained; all unselected categories are treated as background during evaluation. Consistent with~\cite{wu2024point}, raw points are voxelized with a 0.02\,m grid size.

\noindent\textbf{Evaluation Metrics.}
We evaluate performance using mean Intersection-over-Union (mIoU) on three category groups: base classes (mIoU-B), novel classes (mIoU-N), and all classes (mIoU-A). To better capture the balance between base and novel performance, we additionally report the harmonic mean (HM) of mIoU-B and mIoU-N. All results are averaged over five randomly sampled support sets for each evaluation setting.

\noindent\textbf{Implementation Details.}
Following GFS-VL~\cite{an2025generalized}, we build our models on Point Transformer V3 (PTv3)~\cite{wu2024point} as backbone. Training follows a two-stage protocol. We first pretrain on base classes with AdamW (learning rate $6\times10^{-3}$). We then fine-tune in Phase~2 on 1-shot/5-shot support sets for 10\% of the Phase~1 iterations, using a learning rate of $1\times10^{-3}$ on ScanNet200 and $7\times10^{-3}$ on ScanNet++.
 We use HOP-Rep in both phases, and enable HOP-Grad and HOP-Ent only during Phase~2 adaptation. The number of gradient samples $T$ for HOP-Grad is set to 500 to ensure sufficient coverage of base optimization directions. A batch size of 8 is used.
Experiments are run on 8 NVIDIA A100 GPUs. 

\subsection{Main Results}
\label{sec:main_results}
We compare HOP3D with representative GFS-3DS baselines, including PIFS~\cite{cermelli2021prototype}, attMPTI~\cite{zhao2021few}, COSeg~\cite{an2024rethinking}, and GW~\cite{xu2023generalized}, as well as GFS-VL~\cite{an2025generalized}, the state-of-the-art baseline.
The results are quoted from the GFS-VL paper~\cite{an2025generalized}.
A fully supervised model trained with access to both base and novel labels is reported as the upper bound.

\begin{figure}[t]
    \centering
    \includegraphics[width=0.9\linewidth]{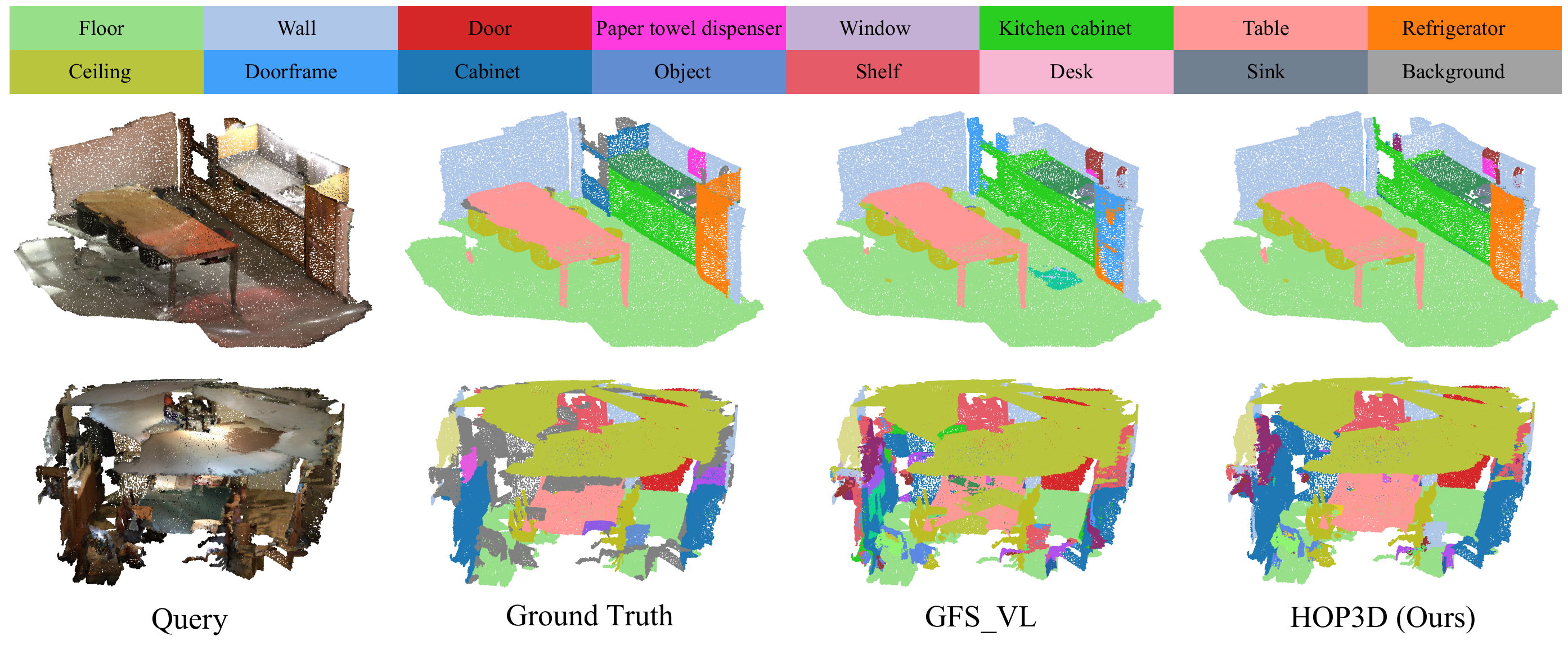}
    \vspace{-6pt}
    \caption{Qualitative comparison between GFS-VL and our HOP3D on ScanNet200. Class color legend is shown at the top. From left to right: query input, ground-truth labels, GFS-VL prediction, and HOP3D prediction.
    }
    \vspace{-10pt}
    \label{fig:main_vis}
\end{figure}

\noindent\textbf{Quantitative results.}
Table~\ref{tab:main-combined} shows that HOP3D outperforms the strongest baseline, GFS-VL, on ScanNet200/++ across most metrics in both 1-shot and 5-shot settings.
On ScanNet200, HOP3D achieves 34.38\% mIoU-N and 45.52\% HM in the 5-shot setting, outperforming GFS-VL by +2.71\% and +2.40\%. The substantial margins over COSeg (+29.17\% mIoU-N and +35.98\% HM) further highlight how hierarchical orthogonalization mitigates prototype collapse—a failure mode observed in prior prototype-refinement approaches.
Under the 1-shot setting on ScanNet200, HOP3D achieves 31.80\% mIoU-N and 43.42\% HM, outperforming GFS-VL by 2.62\% and 2.50\%, respectively, while preserving a highly competitive 68.45\% mIoU-B (compared to GFS-VL's 68.48\%), thereby simultaneously improving novel-class performance and mitigating base-class forgetting under sparse supervision.

On ScanNet++, which features higher scene diversity and larger semantic space, HOP3D continues to maintain strong performance.
In the 5-shot setting, it achieves 23.70\% mIoU-N and 34.34\% HM, surpassing GFS-VL by 2.04\% and 2.52\%. The robust performance on this challenging benchmark suggests that HOP3D scales well with category richness and long-tail distributions.
In the 1-shot scenario, HOP3D attains 19.23\% mIoU-N and 29.32\% HM, achieving highly competitive performance that is comparable to GFS-VL (19.42\% mIoU-N and 29.47\% HM). This is attributed to the entropy-based refinement introduced by HOP-Ent, which mitigates prediction bias and promotes balanced novel-class learning.
Overall, these results demonstrate that HOP3D not only enhances novel-class generalization but also preserves base-class performance by stabilizing the optimization landscape.

\noindent\textbf{Qualitative results.}
Fig.~\ref{fig:main_vis} compares HOP3D with GFS-VL on ScanNet200. 
GFS-VL misclassifies novel objects (e.g., \textit{refrigerator}) as base classes and distorts base-class predictions (e.g., \textit{table} as \textit{ceiling}), whereas HOP3D yields more consistent and accurate segmentation.
These results indicate that HOP3D improves novel-class recognition while preserving base-class segmentation quality.

\subsection{Ablation Study}

\noindent\textbf{Quantitative results.}
We evaluate HOP-Net and HOP-Ent on ScanNet200 under 1-shot, keeping the pseudo-labeling identical to~\cite{an2025generalized}.
As shown in Table~\ref{tab:ablation}, HOP-Rep-only ($\dagger$) and HOP-Grad-only ($\ddagger$) improve mIoU-N/HM by 0.66\%/0.84\% and 0.39\%/0.59\%, respectively, reflecting complementary effects: prototype orthogonalization promotes semantic decoupling, while gradient projection stabilizes few-shot updates by removing base-conflicting directions.
Combining them (full HOP-Net) further boosts mIoU-N/HM to 30.52\%/42.37\% (+1.72\%/+1.98\% over the baseline), indicating additive gains.
We also report a marginal-entropy-only variant ($\S$), which improves mIoU-N/HM by 1.99\%/1.81\% but is below full HOP-Net in HM, suggesting that balancing class frequency alone is insufficient.
Adding full HOP-Ent yields the best trade-off, improving mIoU-N/HM by 3.00\%/3.03\% over the baseline with only a 0.85\% mIoU-B drop vs.\ full HOP-Net.

\begin{figure}[t]
    \centering
    \includegraphics[width=0.9\linewidth]{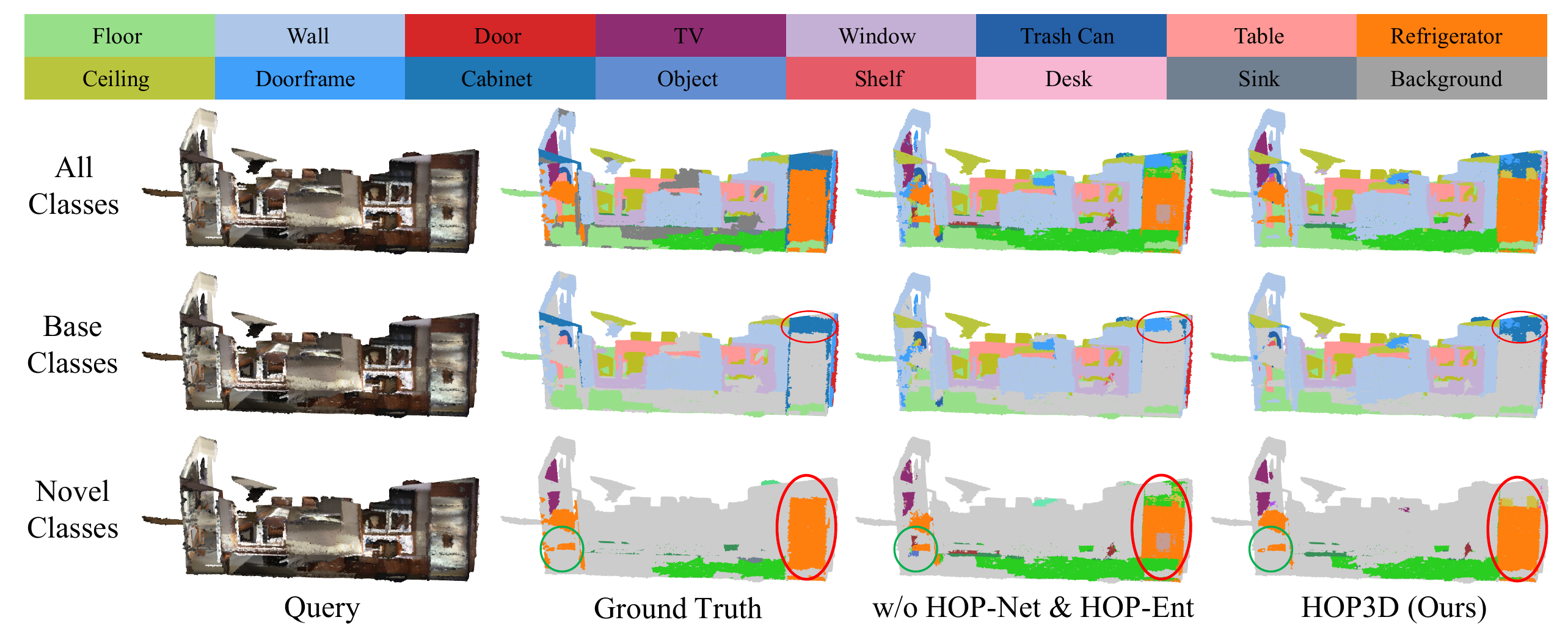}
    \vspace{-6pt}
    \caption{Qualitative results illustrating the effect of HOP-Net and HOP-Ent. From left to right: query, ground truth, prediction without HOP-Net/HOP-Ent, and HOP3D. Circles highlight corrected base–novel misclassifications.}
    \label{fig:ablation_vis}
    \vspace{-10pt}
\end{figure}

\noindent\textbf{Qualitative results.}
Fig.~\ref{fig:ablation_vis} shows that HOP3D corrects typical base--novel confusions (highlighted by circles) compared with the variant without HOP-Net and HOP-Ent.

\begin{table}[t]
\centering
\caption{Ablation on ScanNet200 (1-shot) for HOP-Net and HOP-Ent.}
\vspace{-6pt}
\label{tab:ablation}
\small
\begin{tabular}{c|c|c|c|c|c}
\hline
HOP-Net & HOP-Ent & mIoU-B & mIoU-N & mIoU-A & HM \\
\hline
 & & 67.64 & 28.80 & 36.98 & 40.39 \\
\cmark$^{\dagger}$ & & 68.67 & 29.46 & 37.71 & 41.23 \\ 
\cmark$^{\ddagger}$ & & 68.79 & 29.19 & 37.53 & 40.98 \\ 
\cmark & & \textbf{69.30} & 30.52 & 38.69 & 42.37 \\ 
\cmark & \cmark$^{\S}$ & 67.11 & 30.79 & 38.43 & 42.20 \\ 
\cmark & \cmark & 68.45 & \textbf{31.80} & \textbf{39.52} & \textbf{43.42} \\ 
\hline
\end{tabular}
\vspace{-6pt}
\end{table}

\begin{figure}[t]
    \centering
    \includegraphics[width=0.47\textwidth]{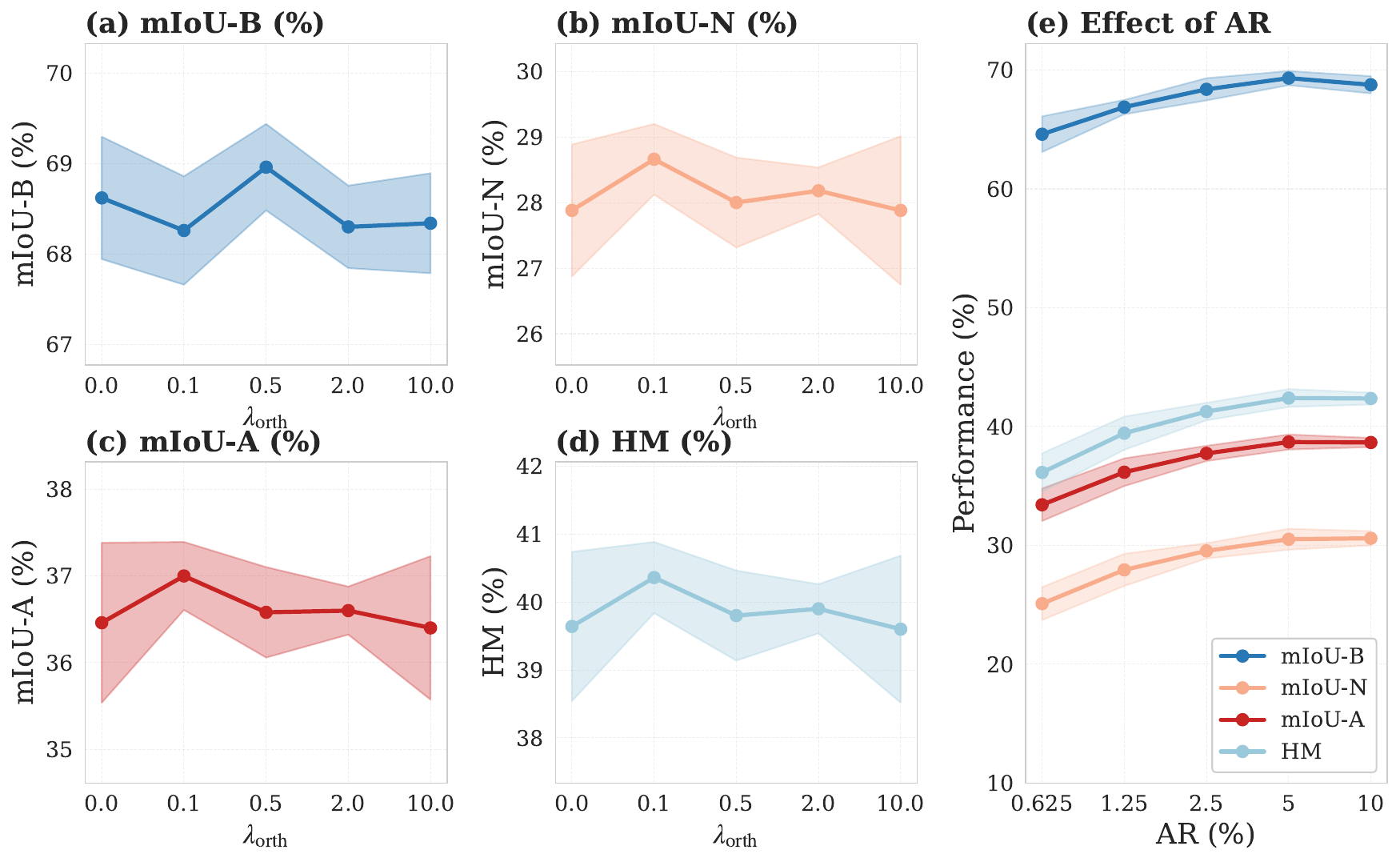}
    \vspace{-6pt}
    \caption{
        HOP-Net ablation. (a)–(d) Impact of $\lambda_{\mathrm{orth}}$. 
        (e) Effect of AR. Shaded areas denote 95\% confidence intervals.
    }
    \vspace{-5pt}
    \label{fig:hog_ablation}
\end{figure}

\begin{figure}[!t]
    \centering
    \includegraphics[width=0.47\textwidth]{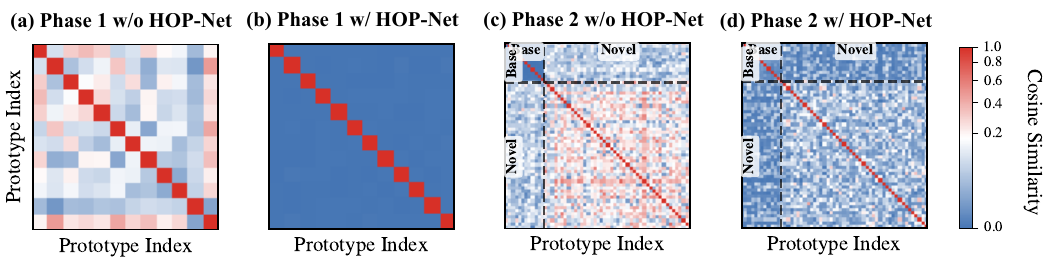}
    \vspace{-8pt}
    \caption{
    Cosine-similarity matrices of $\ell_2$-normalized prototypes (red: higher similarity, blue: lower).
    Phase~1: base prototypes only; Phase~2: joint base+novel prototypes (base first, then novel).
    (a) P1 w/o HOP-Net; (b) P1 w/ HOP-Net; (c) P2 w/o HOP-Net; (d) P2 w/ HOP-Net.
    }
    \vspace{-10pt}
    \label{fig:hog_proto_vis}
\end{figure}

\noindent\textbf{Analysis of HOP-Net.}
HOP-Net introduces hierarchical orthogonalization at both the gradient and prototype levels.
We analyze its behavior by varying the orthogonality weight $\lambda_{\mathrm{orth}}$ and the Phase~2 adaptation ratio (AR).
As shown in Fig.~\ref{fig:hog_ablation}(a)--(d), disabling orthogonalization ($\lambda_{\mathrm{orth}}=0$) hurts mIoU-N and HM, while a small weight improves the base--novel trade-off, with $\lambda_{\mathrm{orth}}=0.1$ giving the best overall balance; we therefore use $\lambda_{\mathrm{orth}}=0.1$ by default.
Fig.~\ref{fig:hog_ablation}(e) further shows that increasing AR consistently benefits both base and novel performance, with even minimal adaptation (0.625\%) yielding clear gains.
As shown in Fig.~\ref{fig:hog_proto_vis}, Phase~2 adaptation without HOP-Net leads to a noticeable surge of off-diagonal prototype similarities, whereas HOP-Net preserves a more diagonal-dominant structure, indicating reduced inter-class redundancy and clearer subspace separation.

\begin{figure}[t]
    \centering
    \includegraphics[width=0.45\textwidth]{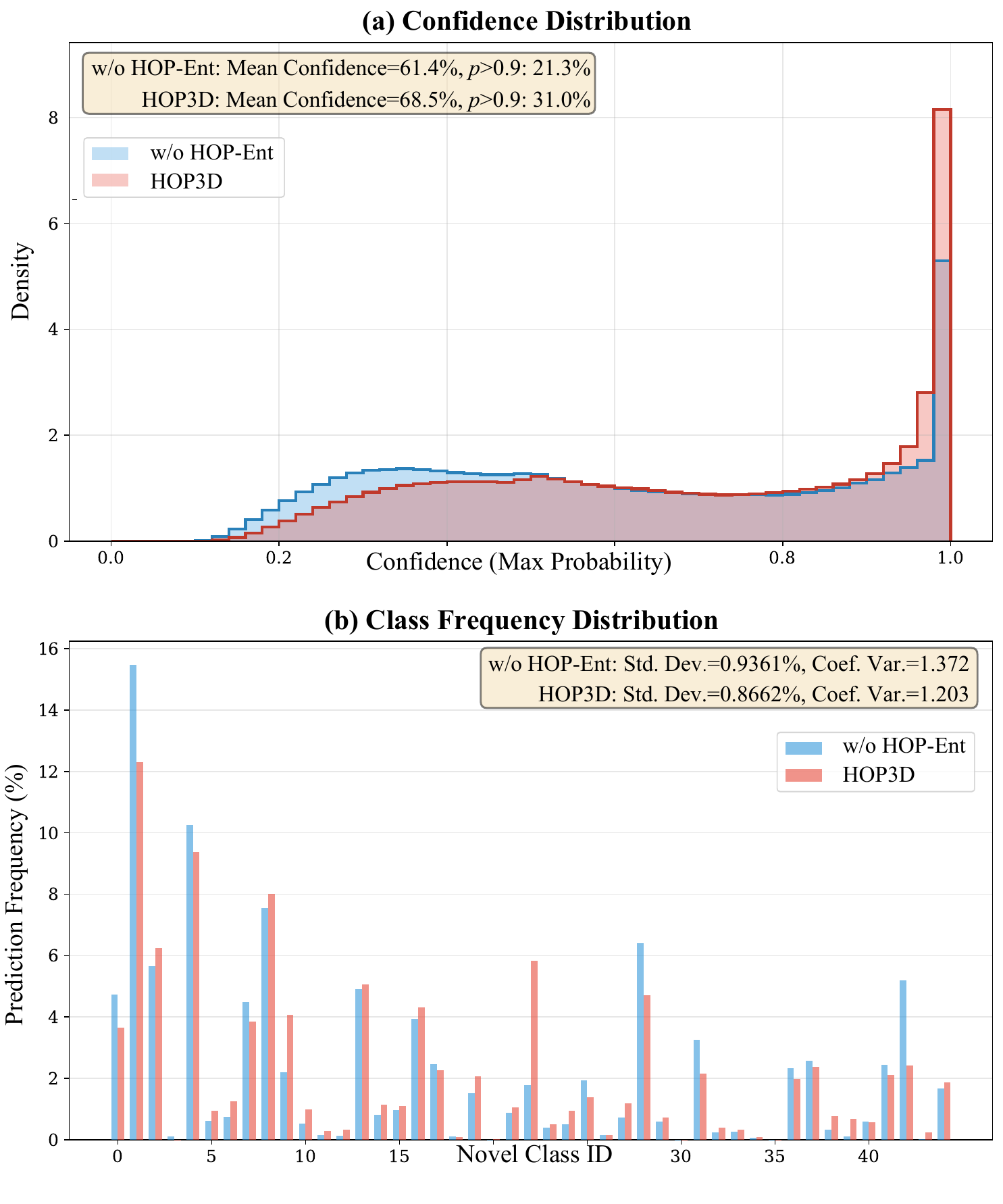}
    \vspace{-10pt}
    \caption{
    HOP-Ent analysis.
    (a) Confidence distribution, where higher mean confidence indicates better prediction certainty.
    (b) Class frequency distribution, where a lower coefficient of variation indicates better class balance.
    HOP-Ent improves both prediction certainty and class balance.
    }
    \vspace{-6pt}
    \label{fig:HOP-Ent_analysis}
\end{figure}

\noindent\textbf{Analysis of HOP-Ent.}
HOP-Ent refines Phase~2 adaptation via conditional and marginal entropy.
As shown in Fig.~\ref{fig:HOP-Ent_analysis}(a), conditional entropy minimization improves prediction certainty, increasing the mean confidence from 61.4\% to 68.5\% and raising the proportion of high-confidence predictions ($p>0.9$) from 21.3\% to 31.0\%.
As shown in Fig.~\ref{fig:HOP-Ent_analysis}(b), marginal entropy maximization improves class balance by reducing the standard deviation of novel-class prediction frequency from 0.9361\% to 0.8662\% and the coefficient of variation from 1.372 to 1.203.
Together, these effects yield more reliable and balanced predictions, improving novel-class performance with minimal impact on base classes, consistent with Table~\ref{tab:ablation}.

\noindent\textbf{Efficiency.}
Measured on the same platform with AR = 10\% (Fig.~\ref{fig:hog_ablation}(e)), HOP3D incurs a 9.7\% training-time overhead over GFS-VL~\cite{an2025generalized}.
Using a smaller AR further reduces the practical overhead, since HOP-Grad is only applied during Phase~2.
Inference cost is unchanged, as HOP-Grad is applied exclusively during training.

\noindent\textbf{Discussion and Limitations.}
Ablations demonstrate that mitigating base--novel interference requires explicitly disentangling optimization dynamics (\emph{how} to adapt) and representation geometry (\emph{what} to adapt) via HOP-Net and HOP-Ent. 
Regarding limitations, our method relies on fixed gradient bases and the pseudo-labeling of~\cite{an2025generalized}; exploring adaptive strategies could further enhance robustness. 
Finally, HOP-Grad introduces a minor training overhead, while inference cost remains unchanged.
\section{Conclusion}
We presented HOP3D, a unified framework for generalized few-shot 3D segmentation that combines hierarchical orthogonal prototype learning with entropy-aware adaptation.
By jointly decoupling optimization and representation (\emph{how} vs.\ \emph{what} to learn), HOP3D is, to the best of our knowledge, the first framework to introduce dual orthogonality into GFS-3DS, achieving a strong balance between base-class retention and novel-class generalization.
HOP-Ent further improves confidence calibration and prediction balance without post-training or test-time procedures.
Extensive experiments on ScanNet200 and ScanNet++ demonstrate state-of-the-art performance (1-shot and 5-shot settings). Future work will explore extensions to cross-modal and open-world 3D scene understanding.
\vspace{8pt}

\section*{Acknowledgment}
This research was supported by the National Key Research and Development Program of China (No.2023YFC3304800).
The computations in this research were performed using the CFFF platform of Fudan University.
A demonstration of this work is deployed on the Cross-Border Trade Payment Intelligent Agent Platform (\href{https://fdueblab.cn/}{fdueblab.cn}).

\bibliographystyle{IEEEbib}
\bibliography{HOP3D}

@string{cvpr = "Proc. IEEE/CVF CVPR"}

@string{iccv   = "Proc. IEEE Int. Conf. Comput. Vis. (ICCV)"}

@string{eccv   = "Proc. Eur. Conf. Comput. Vis. (ECCV)"}

@string{bmvc   = "Proc. Brit. Mach. Vis. Conf. (BMVC)"}

@string{neurips= "Adv. Neural Inf. Process. Syst. (NeurIPS)"}

@string{iclr   = "Int. Conf. Learn. Represent. (ICLR)"}

@string{icassp  = "Proc. ICASSP"}

@inproceedings{qi2017pointnet++,
  title     = {PointNet++: Deep Hierarchical Feature Learning on Point Sets in a Metric Space},
  author    = {Qi, Charles R. and Yi, Li and Su, Hao and others},
  booktitle = neurips,
  year      = {2017}
}

@inproceedings{lai2022stratified,
  title     = {Stratified Transformer for 3D Point Cloud Segmentation},
  author    = {Lai, Xin and Liu, Jianhui and Jiang, Li and others},
  booktitle = cvpr,
  year      = {2022}
}

@article{han2024asgformer,
  title   = {Point Cloud Semantic Segmentation with Adaptive Spatial Structure Graph Transformer},
  author  = {Han, Ting and Chen, Yiping and Ma, Jin and others},
  journal = {International Journal of Applied Earth Observation and Geoinformation},
  year    = {2024}
}

@article{xu2024partwhole,
  title   = {Part-Whole Relational Few-Shot 3D Point Cloud Semantic Segmentation},
  author  = {Xu, S. and Zhang, L. and Jiang, G. and others},
  journal = {Comput. Mater. Con.},
  year    = {2024}
}

@inproceedings{an2025mmfss,
  title     = {Multimodality Helps Few-Shot 3D Point Cloud Semantic Segmentation},
  author    = {An, Zhaochong and Sun, Guolei and Liu, Yun and others},
  booktitle = iclr,
  year      = {2025}
}

@inproceedings{yang2024gfs3d,
  title     = {Generalized Few-Shot 3D Point Cloud Segmentation},
  author    = {Yang, Shuqian and Ding, Henhui and Jiang, Xudong},
  booktitle = {IEEE International Symposium on Circuits and Systems},
  year      = {2024}
}

@inproceedings{kim2023ancl,
  title     = {Achieving a Better Stability--Plasticity Trade-Off via Auxiliary Networks in Continual Learning},
  author    = {Kim, Sanghwan and Noci, Lorenzo and Orvieto, Antonio and others},
  booktitle = cvpr,
  year      = {2023}
}

@inproceedings{thomas2019KPConv,
  title     = {KPConv: Flexible and Deformable Convolution for Point Clouds},
  author    = {Thomas, Hugues and Qi, Charles R. and Deschaud, Jean-Emmanuel and others},
  booktitle = iccv,
  year      = {2019}
}

@article{kirkpatrick2017ewc,
  title   = {Overcoming Catastrophic Forgetting in Neural Networks},
  author  = {Kirkpatrick, James and Pascanu, Razvan and Rabinowitz, Neil and others},
  journal = {Proceedings of the National Academy of Sciences},
  year    = {2017}
}

@article{parisi2019continual,
  title   = {Continual Lifelong Learning with Neural Networks: A Review},
  author  = {Parisi, German I. and Kemker, Ronald and Part, Jose L. and others},
  journal = {Neural Networks},
  year    = {2019}
}

@article{sun2024review,
  title   = {A Review of Point Cloud Segmentation for Understanding 3D Indoor Scenes},
  author  = {Sun, Yuliang and Zhang, Xudong and Miao, Yongwei},
  journal = {Visual Intelligence},
  year    = {2024}
}

@article{wang2024survey,
  title   = {A Survey on Weakly Supervised 3D Point Cloud Semantic Segmentation},
  author  = {Wang, Jingyi and Liu, Yu and Tan, Hanlin and others},
  journal = {IET Computer Vision},
  year    = {2024}
}

@inproceedings{schult2023mask3d,
  title     = {Mask3D: Mask Transformer for 3D Semantic Instance Segmentation},
  author    = {Schult, Jonas and Engelmann, Francis and Hermans, Alexander and others},
  booktitle = {IEEE International Conference on Robotics and Automation},
  year      = {2023}
}

@inproceedings{kolodiazhnyi2024oneformer3d,
  title     = {OneFormer3D: One Transformer for Unified Point Cloud Segmentation},
  author    = {Kolodiazhnyi, Maxim and Vorontsova, Anna and Konushin, Anton and others},
  booktitle = cvpr,
  year      = {2024}
}

@inproceedings{hajimiri2023strong,
  title     = {A Strong Baseline for Generalized Few-Shot Semantic Segmentation},
  author    = {Hajimiri, Sina and Boudiaf, Malik and Ben Ayed, Ismail and others},
  booktitle = cvpr,
  year      = {2023}
}

@inproceedings{farajtabar2020orthogonal,
  title     = {Orthogonal Gradient Descent for Continual Learning},
  author    = {Farajtabar, Mehrdad and Azizan, Navid and Mott, Alex and others},
  booktitle = {International Conference on Artificial Intelligence and Statistics},
  year      = {2020}
}

@inproceedings{liu2023learning,
  title     = {Learning Orthogonal Prototypes for Generalized Few-Shot Semantic Segmentation},
  author    = {Liu, Sun-Ao and Zhang, Yiheng and Qiu, Zhaofan and others},
  booktitle = cvpr,
  year      = {2023}
}

@inproceedings{an2025generalized,
  title     = {Generalized Few-Shot 3D Point Cloud Segmentation with Vision-Language Model},
  author    = {An, Zhaochong and Sun, Guolei and Liu, Yun and others},
  booktitle = cvpr,
  year      = {2025}
}

@inproceedings{zhao2021few,
  title     = {Few-Shot 3D Point Cloud Semantic Segmentation},
  author    = {Zhao, Na and Chua, Tat-Seng and Lee, Gim Hee},
  booktitle = cvpr,
  year      = {2021}
}

@inproceedings{an2024rethinking,
  title     = {Rethinking Few-Shot 3D Point Cloud Semantic Segmentation},
  author    = {An, Zhaochong and Sun, Guolei and Liu, Yun and others},
  booktitle = cvpr,
  year      = {2024}
}

@inproceedings{rozenberszki2022language,
  title     = {Language-Grounded Indoor 3D Semantic Segmentation in the Wild},
  author    = {Rozenberszki, David and Litany, Or and Dai, Angela},
  booktitle = eccv,
  year      = {2022}
}

@inproceedings{yeshwanth2023scannet++,
  title     = {ScanNet++: A High-Fidelity Dataset of 3D Indoor Scenes},
  author    = {Yeshwanth, Chandan and Liu, Yueh-Cheng and Nissner, Matthias and others},
  booktitle = iccv,
  year      = {2023}
}

@inproceedings{dai2017scannet,
  title     = {ScanNet: Richly-Annotated 3D Reconstructions of Indoor Scenes},
  author    = {Dai, Angela and Chang, Angel X. and Savva, Manolis and others},
  booktitle = cvpr,
  year      = {2017}
}

@inproceedings{xu2023generalized,
  title     = {Generalized Few-Shot Point Cloud Segmentation via Geometric Words},
  author    = {Xu, Yating and Hu, Conghui and Zhao, Na and others},
  booktitle = iccv,
  year      = {2023}
}

@inproceedings{wu2024point,
  title     = {Point Transformer V3: Simpler, Faster, Stronger},
  author    = {Wu, Xiaoyang and Jiang, Li and Wang, Peng-Shuai and others},
  booktitle = cvpr,
  year      = {2024}
}

@inproceedings{cermelli2021prototype,
  title     = {Prototype-Based Incremental Few-Shot Segmentation},
  author    = {Cermelli, Fabio and Mancini, Massimiliano and Xian, Yongqin and others},
  booktitle = bmvc,
  year      = {2021}
}

@article{he2023prototype,
	title        = {Prototype Adaption and Projection for Few- and Zero-Shot 3D Point Cloud Semantic Segmentation},
	author = {He, Shuting and Jiang, Xudong and Jiang, Wei and others},
	year         = 2023,
	journal      = {{IEEE} Trans. Image Process.},
	volume       = 32,
	pages        = {3199--3211},
	doi          = {10.1109/TIP.2023.3279660},
	url          = {https://doi.org/10.1109/TIP.2023.3279660},
	bibsource    = {dblp computer science bibliography, https://dblp.org}
}

@inproceedings{chowdhury2022fscil3d,
  title     = {Few-Shot Class-Incremental Learning for 3D Point Cloud Objects},
  author    = {Chowdhury, Townim and Cheraghian, Ali and Ramasinghe, Sameera and others},
  booktitle = eccv,
  year      = {2022}
}

@inproceedings{bansal2018orthogonal,
  title     = {Can We Gain More from Orthogonality Regularizations in Training Deep CNNs?},
  author    = {Bansal, Naman and Chen, Xiaohan and Wang, Zhangyang and others},
  booktitle = neurips,
  year      = {2018}
}

@inproceedings{cogswell2015decov,
  title     = {Reducing Overfitting in Deep Networks by Decorrelating Representations},
  author    = {Cogswell, Michael and Ahmed, Faruk and Girshick, Ross and others},
  booktitle = iclr,
  year      = {2016}
}

@inproceedings{zhao2026upl,
  title     = {Uncertainty-aware Prototype Learning with Variational Inference for Few-shot Point Cloud Segmentation},
  author    = {Zhao, Yifei and Zhao, Fanyu and Li, Yinsheng},
  booktitle = icassp,
  year      = {2026}
}

@article{wei2025few,
  title   = {Few-Shot 3D Point Cloud Segmentation via Relation Consistency-Guided Heterogeneous Prototypes},
  author  = {Wei, Lili and Lang, Congyan and Xu, Zheming and others},
  journal = {IEEE TMM},
  year    = {2025}
}

@string{cvpr   = "Proc. IEEE/CVF Conf. Comput. Vis. Pattern Recognit. (CVPR)"}

\end{document}